\documentclass[final]{cvpr}

\usepackage{times}
\usepackage{epsfig}
\usepackage{graphicx}
\usepackage{amsmath}
\usepackage{amssymb}
\usepackage{subfigure}
\usepackage{algorithm}
\usepackage{algorithmic}
\usepackage{adjustbox,lipsum}
\usepackage[table,dvipsnames]{xcolor}
\usepackage{multirow}
\usepackage{makecell}
\usepackage{caption}
\usepackage{color}
\usepackage{subfiles}

\usepackage{pifont}
\newcommand{\cmark}{\ding{51}}%
\usepackage[pagebackref=true,breaklinks=true,colorlinks,bookmarks=false]{hyperref}

\def\cvprPaperID{1510} 
\def\confYear{CVPR 2021}

\begin{document}

\title{Picasso: A CUDA-based Library for Deep Learning over 3D Meshes\vspace{-3mm}} 

\author{Huan Lei\hspace{6mm}Naveed Akhtar\hspace{6mm}Ajmal Mian\\
The University of Western Australia\\
{\tt\small huan.lei@research.uwa.edu.au, \{naveed.akhtar,ajmal.mian\}@uwa.edu.au}\\
{\small Code:~\tt \href{https://github.com/hlei-ziyan/Picasso}{https://github.com/hlei-ziyan/Picasso}}
\vspace{-6mm}}
\maketitle

\begin{abstract}
\vspace{-2mm}
We present Picasso, a CUDA-based library comprising novel modules for deep learning over complex real-world 3D meshes.
Hierarchical neural architectures have proved effective in multi-scale feature extraction which signifies the need for fast mesh decimation. 
However, existing methods rely on CPU-based implementations to obtain multi-resolution meshes.
We design GPU-accelerated mesh decimation to facilitate network resolution reduction efficiently on-the-fly. Pooling and unpooling
 modules are defined on the vertex clusters gathered during decimation. For feature learning over meshes, Picasso contains three types of novel convolutions namely, facet2vertex, vertex2facet, and facet2facet convolution. Hence, it treats a mesh as a geometric structure comprising vertices and facets, rather than a spatial graph with edges as previous methods do.
Picasso also incorporates a fuzzy mechanism in its filters for robustness to mesh sampling (vertex density).
It exploits Gaussian mixtures to define fuzzy coefficients for the facet2vertex convolution, and barycentric
interpolation to define the coefficients for the remaining two convolutions. In this release, we  demonstrate the effectiveness of the  proposed modules with competitive segmentation results on S3DIS. 
The library will be made public through \href{https://github.com/hlei-ziyan/Picasso}{github}.
\end{abstract}

\vspace{-6mm}
\section{Introduction}
\vspace{-2mm}
Data in computer vision vary commonly from homogeneous format in 2D projective space (e.g.~images, videos) to heterogeneous format in 3D Euclidean space (e.g.~point clouds, meshes). The success of convolutional feature learning on homogeneous data \cite{he2016deep,krizhevsky2012imagenet,liu2016ssd,long2015fully,redmon2016you,ren2015faster,ronneberger2015u,simonyan2014very} has sparked research interest in geometric deep learning \cite{bronstein2017geometric,bruna2013spectral,defferrard2016convolutional,kipf2017semi,simonovsky2017dynamic}, which aims for equally effective feature learning on heterogeneous data. Due to the rise of 
autonomous driving and robotics, 3D deep learning has now become an important branch of the geometric research direction.
Compared to 3D point clouds, 3D meshes convey richer geometric information about the object surface and topology. 
Yet, the heterogeneous facet shapes and sizes combined with unstructured vertex locations make its adaption to deep learning more difficult as compared to point clouds. 
This is why most approaches address real-world 3D scene understanding via convolutions on point clouds \cite{lei2019octree,lei2020seggcn,lei2020spherical,li2018pointcnn,qi2017pointnet,qi2017pointnetplusplus}. 
However, point clouds still lack in preserving the structural details that are easily represented by meshes.

There are a few works that learn features from meshes, but they are largely constrained to \textit{shape analysis} on small synthetic models \cite{boscaini2016learning,hanocka2019meshcnn,masci2015geodesic,monti2017geometric,ranjan2018generating,xie2015deepshape}. 
These methods either apply convolutions throughout the network to a single mesh resolution (the input), or exploit inefficient CPU-based algorithms to decimate the mesh \cite{garland1999quadric,garland1997surface,rossignac1993multi,zhou2018open3d}. 
However, non-hierarchical network configurations and slow network coarsening are both problematic while dealing with real-world meshes because of their large-scale nature. 
This calls for mesh simplification methods that are fast and amenable to deep learning for the real-world applications.

\begin{figure*}[!t]
    \centering
    \begin{tabular}{cccc}
    \hspace{-2mm}
    \includegraphics[width=0.24\textwidth]{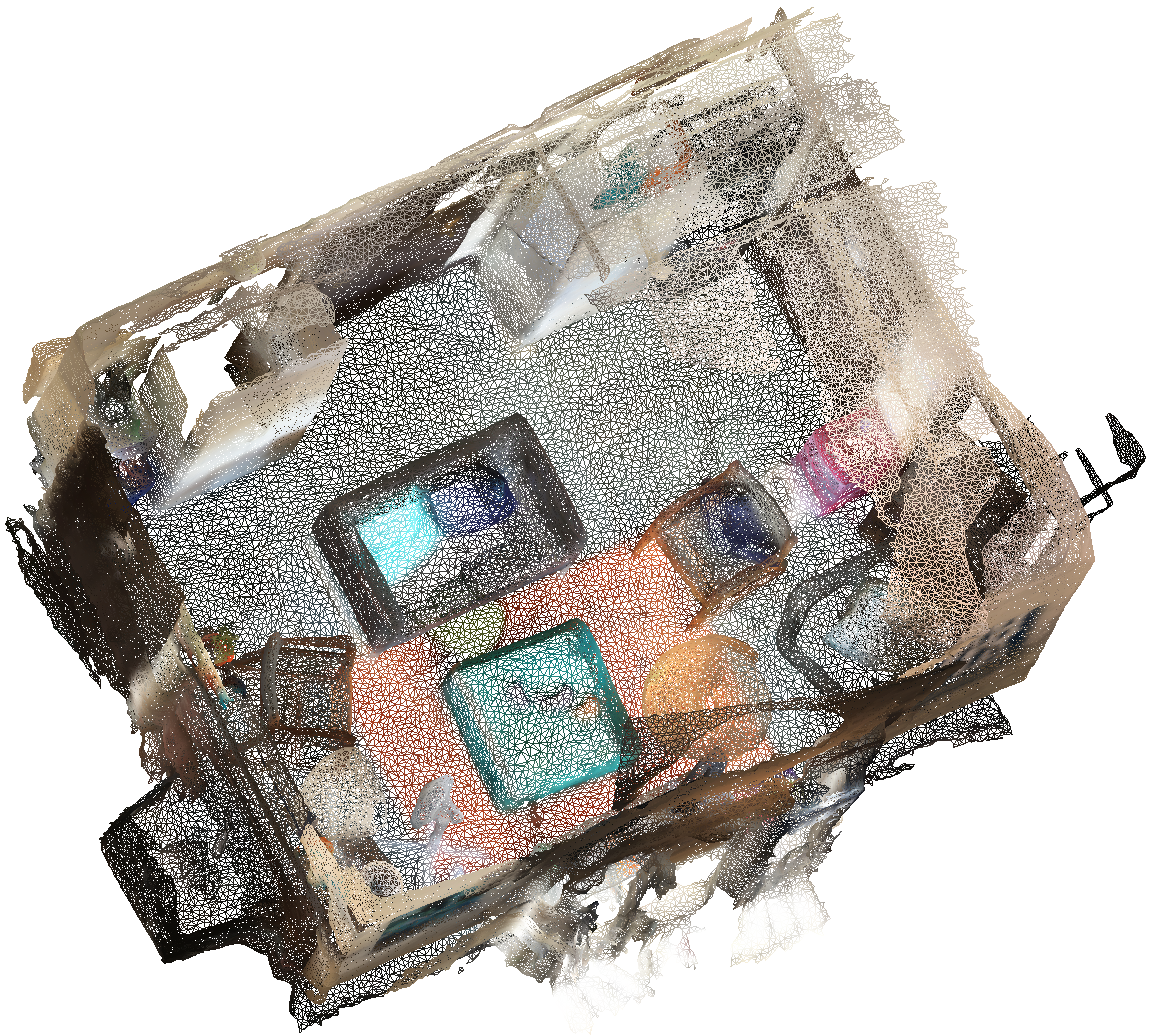}&
    \hspace{-3mm}
    \includegraphics[width=0.24\textwidth]{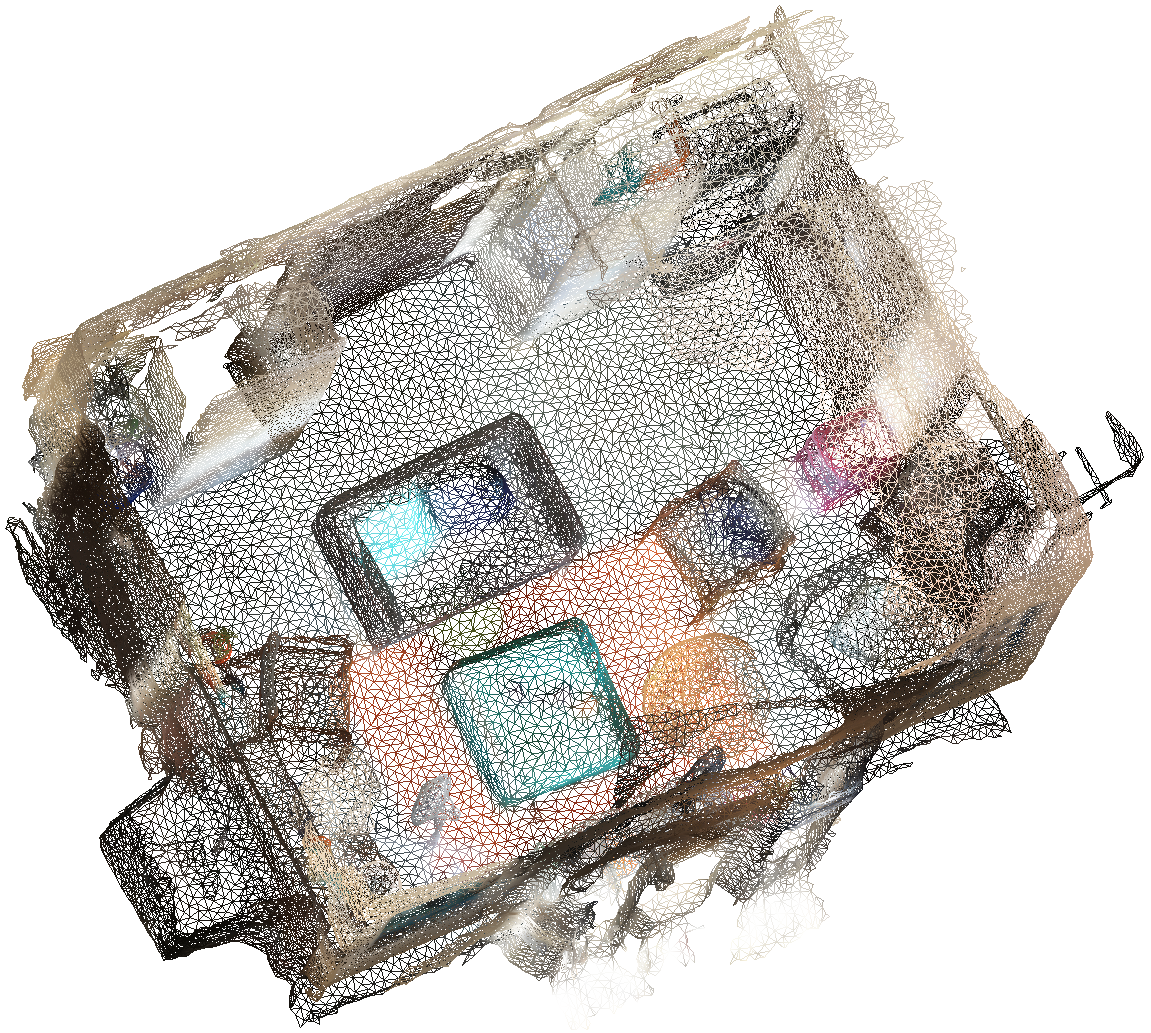} &
    \hspace{-3mm}
    \includegraphics[width=0.24\textwidth]{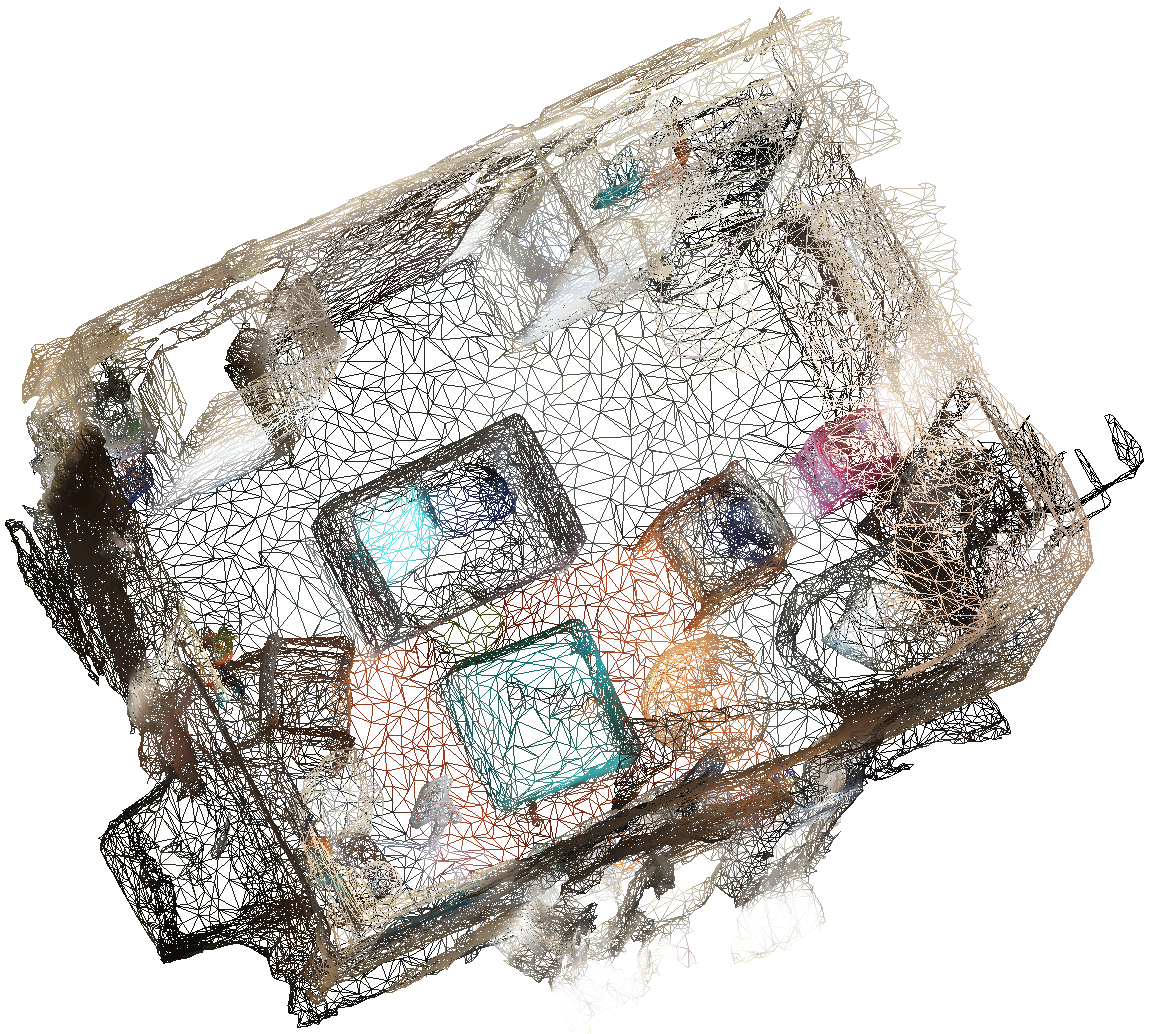}&
    \hspace{-3mm}
    \includegraphics[width=0.24\textwidth]{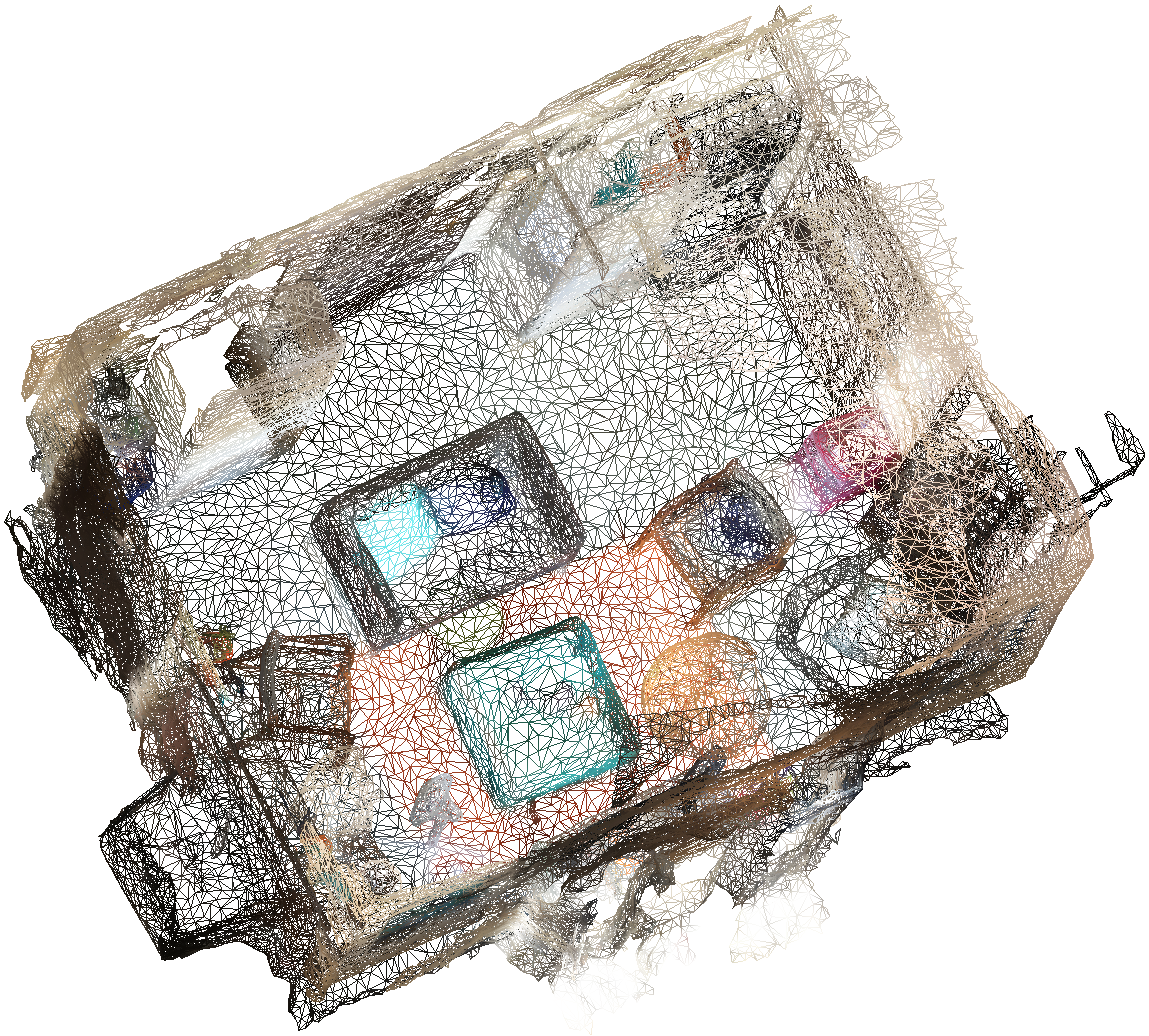} \\
    \hspace{-10mm}(a) The input mesh & (b) VC: $177$~ms& (c) QEM: $2160$~ms & (d) Ours: ${\bf 65}$~ms  \\
    \end{tabular}
    \vspace{-2mm}
\caption{Runtime comparison of mesh simplification. The input mesh consists of $115,114$ vertices and $231,293$ facets. Different methods simplify it to similar mesh sizes. The number of vertices and facets in the decimated meshs are $41,483$ and $85,424$ for vertex clustering (VC)~\cite{rossignac1993multi}, $41,133$ and $83,050$ for the quadric error metrics (QEM)~\cite{garland1999quadric,garland1997surface}; and $41,449$ and $83,051$ for our GPU-accelerated algorithm. The runtime of VC, QEM, and our method are respectively $177$~ms, $2160$~ms, and $65$~ms.  For VC and QEM, we utilize their popular implementations in Open3D~\cite{zhou2018open3d}. For better visualization, we show wireframes of the meshes only. Best viewed in color and enlarged.}
    \label{fig:decimate_runtimes}
    \vspace{-3mm}
\end{figure*}
We present a GPU-accelerated mesh simplification algorithm to facilitate the exploration of hierarchical architectures on meshes.
The proposed method is not only fast in decimating small-scale watertight meshes from CAD modelling \cite{Bogo:CVPR:2014,chang2015shapenet,cosmo2016shrec,lian2011shape}, but is also efficient in simplifying large-scale unstructured real-wold meshes \cite{armeni20163d,Matterport3D,dai2017scannet,scenenn-3dv16}. 
We perform all computations in parallel on GPU, except for the grouping of vertex pairs to be contracted.
Meanwhile, to increase its compatibility with modern deep learning modules such as normalization \cite{ba2016layer,ioffe2015batch, ulyanov2016instance, wu2018group}, we contract vertex clusters by controlling the desired vertex size of the decimated mesh. This also advances mesh-based modules to be exploited in conjunction with the various point cloud based modules \cite{schult2020dualconvmesh}.
Our algorithm is able to reduce the number of mesh vertices by half in each iteration. Figure \ref{fig:decimate_runtimes} compares the runtime of our method with two well-founded decimation methods, VC~\cite{rossignac1993multi} and QEM~\cite{garland1997surface}. Notice that our method is $30\times$ faster than QEM. 
During the simplification, 
we record all vertex clustering information into a 1-D tensor. Based on this tensor, we also 
define max, average and weighted poolings, as well as unpooling.

Earlier attempts for convolution on meshes \cite{boscaini2016learning,masci2015geodesic,monti2017geometric} explored local patch operators in hand-crafted coordinate systems. The development of spatial graph convolutions has led recent methods \cite{hanocka2019meshcnn,ranjan2018generating,schult2020dualconvmesh} to predominantly consider (\textit{triangular}) mesh as a special graph and convolve features of each vertex from their geodesic $k$-ring 
neighborhood. 

In contrast, we 
study mesh as a set of vertices and facets, following its natural geometric structure. 
To learn features of each vertex, we aggregate their context 
information from the adjacent facets. We refer to the resulting operation as  \textit{facet2vertex} convolution. 
Lei \emph{et al.} \cite{lei2020seggcn}
showed that fuzzy mechanism makes the convolutional network robust to point density variation.  Hence, we further exploit fuzzy coefficients in the facet2vertex convolution. 
Due to the fact that facet normals are  distributed strictly on the surface of a unit sphere, i.e.~${\bf S}^3=\{{\bf x}\in{\bf R}^3: \|{\bf x}\|=1\}$, we associate learnable filter weights to  Gaussian mixtures defined on ${\bf S}^3$. The parameters of the Gaussian components can optionally be kept fixed or trainable within the network in our library. 
On the other hand, to learn features of the mesh facets, we introduce  \textit{vertex2facet} and \textit{facet2facet} convolutions.
The former propagates features from the vertices of a facet to the facet itself, while the latter is applied when facets of the input mesh are rendered with textures. 
We incorporate fuzziness into these two convolutions using barycentric interpolation. 
The three proposed convolutions altogether enable flexible vertex-wise and facet-wise feature learning on the mesh.

We provide CUDA implementations for all the above mentioned modules and organize them into a self-contained library, named Picasso\footnote{Paying homage to Pablo Picasso for cubism in paintings.}, to facilitate deep learning over the unstructured real-world 3D meshes. We note that meshes and point clouds are tightly bonded together, and it is more desirable to extract features from the two data modalities cooperatively rather than individually or competitively. DCM-Net \cite{schult2020dualconvmesh} also validates this argument. For this reason, we additionally incorporate all the point cloud modules from Lei \emph{et al.} \cite{lei2020seggcn,lei2020spherical} in our library (with author permission). 
In this maiden release of our library, we demonstrate promise of its proposed modules with competitive segmentation results on S3DIS \cite{armeni20163d}. The segmentation network is abbreviated as PicassoNet for consistency. 
We summarize the main contributions of our work below:
\begin{itemize}
 \vspace{-1mm}
\item We present a fast GPU-accelerated 3D mesh decimation technique to reduce mesh resolution on-the-fly. A public implementation with complementary CUDA-based pooling and unpooling operations is provided.  

\vspace{-1mm}
 \item We propose three novel convolution modules, i.e. \textit{facet2vertex}, \textit{vertex2facet}, \textit{facet2facet}, to alternatively learn vertex-wise and facet-wise features on a mesh. Diverging from  existing  methods, we do not rely on restrictive treatment of mesh as an undirected graph with edges. Instead, it's a geometric structure composed of vertices and facets for our modules, which is also a more conducive  representation for the digital devices.
 
 \vspace{-1mm}
\item With this paper, we release Picasso --- a self-contained library for deep learning over  unstructured real-world 3D meshes, along with  synthetic watertight meshes. The provided anonymous \href{https://anonymous.4open.science/r/e3278231-2703-4a59-b10c-1d33c37333a7/}{github link} will be made public for the broader research community.
\end{itemize}

\vspace{-3mm}
\section{Related Work}
\vspace{-1mm}
\noindent\textbf{Convolutions on point clouds:} 
3D-CNNs \cite{EngelckeICRA2017,graham20183d,maturana2015voxnet,riegler2017octnet,wu20153d} are the most intuitive solutions of transferring CNNs from images to point clouds. A few methods also explore similar regular-grid kernels in a transformed data domain \cite{su2018splatnet,tatarchenko2018tangent}. The permutation invariant networks exploit multi-layer perceptrons (MLPs) and max-pooling to learn features from point clouds \cite{klokov2017escape,li2018so,qi2017pointnet,qi2017pointnetplusplus,rethage2018fully,shen2018mining,wu2019pointconv}. They demonstrate the effectiveness of taking point coordinates $xyz$ as input features.
Graph-based networks allow convolutions to be performed in either spectral or spatial domain. 
However, the mandatory alignment of different graph Laplacians makes the application of spectral graph convolutions to point clouds more difficult than spatial graph convolutions \cite{yi2017syncspeccnn}. As a pioneering work, ECC~\cite{simonovsky2017dynamic} exploited dynamic filters \cite{de2016dynamic} to analyze point clouds with the spatial graph convolutions. Subsequent works explored more effective filter or kernel parameterizations \cite{groh2018flex,li2018pointcnn,wang2019attention,wu2019pointconv,xu2018spidercnn}.
The recent discrete kernels  \cite{lei2019octree,lei2020seggcn,lei2020spherical,thomas2019KPConv} 
are appealing alternatives to those dynamic methods as they avoid the dependence of filter generations within the network. 
KPConv~\cite{thomas2019KPConv} reported more competitive results, while the spherical kernels \cite{lei2020seggcn,lei2020spherical} are more memory and runtime efficient. 

\noindent\textbf{Convolution on meshes:}
In nascency of this direction, 
researchers generally performed convolutions over local patches defined in a hand-crafted coordinate system \cite{boscaini2016learning,masci2015geodesic,monti2017geometric}. The coordinate system could  either be established by geodesic level sets \cite{masci2015geodesic}
or surface normals and principle curvatures \cite{boscaini2016learning,monti2017geometric}.
In FeaStNet \cite{verma2018feastnet}, Verma \emph{et al.} capitalized on a learnable
mapping between filter weights and graph neighborhoods to replace those hand-crafted local patches.
TextureNet \cite{huang2019texturenet} takes surface meshes with high-resolution facet textures as input, and explores a 4-RoSy field to parameterize the mesh into local planar patches such that standard CNNs \cite{krizhevsky2012imagenet} are applicable. 
Ranjan \emph{et al.}~\cite{defferrard2016convolutional} proposed to learn human facial expressions with hierarchical mesh-based networks, using the spectral graph convolutions.  
Schult \emph{et al.}~\cite{schult2020dualconvmesh} proposed to extract features from both meshes and point clouds simultaneously. Similar to \cite{li2018pointcnn,simonovsky2017dynamic,wang2018dynamic,wu2019pointconv}, they still conduct convolution using dynamically generated filters. Whereas most methods learn vertex-wise features,
MeshCNN \cite{hanocka2019meshcnn} defines convolution to learn edge-wise features on a mesh.

Generally, previous methods treat mesh as an edge-based graph and employ geodesic convolutions over it. In this paper, we explore convolutions on the mesh following its own geometric structure, i.e.~vertices and facets. To promote this more natural perspective,  we also provide computation and memory optimized CUDA implementations for forward and backward propagations of all the convolutions we propose. 

\noindent\textbf{Mesh decimation:} 
Hierarchical networks allow convolutions to be applied on an increasing receptive fields of the input data. To create such hierarchical architectures on point clouds, researchers usually exploit random point
sampling or farthest point sampling (FPS)  \cite{lei2020spherical,qi2017pointnetplusplus,wu2019pointconv}. However, because of their inability to track vertex connections, they are not applicable to mesh processing.
Fortunately, mesh simplification is a well-studied topic in the graphics community. There are methods available that can be used to decimate a mesh
\cite{garland1999quadric,garland1997surface,rossignac1993multi}. For example, Ranjan \emph{et al.}~\cite{ranjan2018generating} used the quadric error metrics \cite{garland1997surface} to simplify their synthetic facial meshes. Schult \emph{et al.} \cite{schult2020dualconvmesh}
explored vertex clustering \cite{rossignac1993multi} and quadric error metrics \cite{garland1999quadric,garland1997surface} to simplify their indoor room meshes. In particular, they made use of the simplification functions provided by Open3D \cite{zhou2018open3d} - a popular library for 3D geometry processing in Python.
However, the implementations in Open3D are CPU-based, which are not amenable to deep learning. 

In this work, we also introduce a fast mesh decimation method based on the algorithm of Garland \emph{et al.} \cite{garland1999quadric,garland1997surface}.
Their method simplifies a mesh through iterative contractions of vertex pairs, and
demands that the vertex pair contribution to quadric error be determined after each iteration. This progressive strategy makes the algorithm impossible for parallel deployment on GPUs. In comparison, we sort all the vertex pairs by their quadric errors only once, and group them into isolated vertex clusters. All other computations in our method are mutually independent and can be accelerated via parallel GPU computing. We represent the vertex cluster information as a 1-D tensor to facilitate the pooling and unpooling operations.

\vspace{-2mm}
\section{GPU-Accelerated Mesh Decimation}
\vspace{-1mm}
To explore flexible deep neural networks for meshes, there has been an increasing demand for a fast mesh decimation method in the 3D community. The quadric error simplification \cite{garland1997surface} performs iterative contraction of vertex pairs until the desired number of facets is obtained. This method is known to be effective for simplifying meshes while retaining the quality. 
However, it is not suited to parallel computing due to the implicit dependencies between its iterative contractions. Considering its high  relevance, we provide the quadric error method as Algorithm~\ref{alg:QEM_garland}. 
The iterative dependency of the method is clear from \textit{lines 7--13} of the algorithm. 

\begin{figure}[!t]
    \centering
    \includegraphics[width=0.46\textwidth]{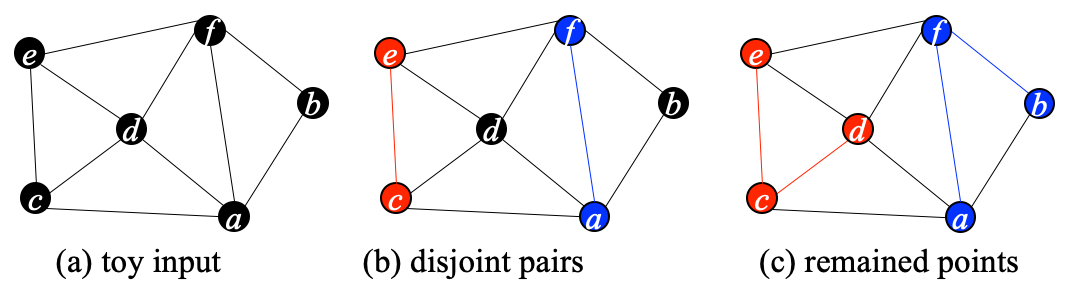}
    \vspace{-3mm}
\caption{Illustration of edge-based vertex clustering. (a) shows an input mesh with 10 vertex pairs. (b) initializes the disjoint vertex pairs $(c,e),~(a,f)$ as clusters $\{c,e\},~\{a,f\}$ - shown by red and blue. (c) groups the remaining points $d,~b$ to the vertex clusters they each connect to, resulting in the final vertex clusters $\{c,d,e\}$ and $\{a,b,f\}$.} 
\label{fig:vertex_cluster}
\vspace{-5mm}
\end{figure}

We extend \cite{garland1997surface} and propose a fast mesh decimation method that exploits the parallel computing power of a GPU. The main difference is that we do not perform iterative contractions any more. Instead, we group the vertices into multiple isolated clusters based on their connections. Figure~\ref{fig:vertex_cluster} provides a toy example to illustrate the clustering process.
During clustering, we control the expected vertex number rather than the number of facets or edges. 
Since the contractions of isolated clusters are independent of each other, they can be executed on a GPU in parallel. 
More specifically, we initialize the candidates of vertex pairs to be contracted using existing mesh edges only. The vertex clusters are established with disjoint vertex pairs in the candidates, before which we sort the candidates in an ascending order by their quadric cost\footnote{Shuffling strategy is inserted into the sorting of quadrics for variations of the decimated mesh at different epochs.}.
See Algorithm \ref{alg:Mesh_decimate_ours} for our method. We note that our core algorithm reduces the vertex number roughly by a half, per-iteration. Yet, we can still handle arbitrary number of output vertices. It is achieved by running the core algorithm multiple times. 
For clarity, we present Algorithm~\ref{alg:Mesh_decimate_ours} with a single mesh taken as the input. In our library, the decimation function can take mini-batches of concatenated meshes as input. This improves its compatibility with deep learning, e.g.~batch normalization~\cite{ioffe2015batch} is easily applicable. The vertex clustering operations (\textit{lines 3--11} in Algorithm \ref{alg:Mesh_decimate_ours}) is currently  executed on CPU, that helps in deploying heavy computations to GPU.
We note that consistency checks and penalties
are excluded in our method in favor of better runtime efficiency.
For continuity, we delegate the details on complementary output parameters \textit{replace} and \textit{mapping} to the source code itself. 


\begin{algorithm}[!t]
 \caption{The original quadric error simplification}
 \label{alg:QEM_garland}
 \begin{algorithmic}[1]
 \renewcommand{\algorithmicrequire}{\textbf{Input:}}
 \renewcommand{\algorithmicensure}{\textbf{Output:}}
 \REQUIRE A triangular mesh $\mathcal{T}^i=(\mathcal{V}^i,\mathcal{F}^i)$, the number of output facets $M$, threshold $\tau$.
 \ENSURE A decimated mesh ${\mathcal{T}^o}=(\mathcal{V}^o,\mathcal{F}^o)$.
 \STATE  {select any pair $({\bf v}_i$, ${\bf v}_j)$ that is an edge or $\|{\bf v}_i-{\bf v}_j\|<\tau$ as candidates} 
\STATE {compute the quadric of each vertex pair}
\FOR{each candidate pair $({\bf v}_i$, ${\bf v}_j)$}
\STATE (b) determine the target position $\bar{\bf v}$ of contraction
\STATE (c) apply consistency checks and penalties
\ENDFOR
\REPEAT
\IF{the pair (${\bf v}_i$, ${\bf v}_j$) has the least quadric cost $Q$}
\STATE (b) perform contraction $({\bf v}_i, {\bf v}_j) \rightarrow \bar{\bf v}$
\STATE (c) update $Q_i = Q_i + Q_j$
\STATE (d) for each affected pair (${\bf v}_i$, ${\bf v}_k$), recompute its target position and cost as in steps 3-6
\ENDIF
\UNTIL{$|\mathcal{F}^o|<M$}
\STATE return
 \end{algorithmic}
 \end{algorithm}

\vspace{1mm}
\noindent \textbf{Pooling and unpooling:}
We record the vertex clustering information as well as the degenerated vertex information into different 1-D tensors. The sizes of the two tensors are both identical to the input vertex number.
They largely facilitate the pooling and unpooling operations. 
Given the vertex clusters as neighborhood, different pooling operations can be defined, such as sum$(\cdot)$, average$(\cdot)$, max$(\cdot)$ and weighted$(\cdot)$. We provide average$(\cdot)$, max$(\cdot)$, weighted$(\cdot)$ poolings in the library. For unpooling, we simply interpolate features of all vertices in a cluster based on features of the representative vertex in the cluster. 

\begin{algorithm}[!t]
\caption{Fast GPU-accelerated mesh simplification}
\label{alg:Mesh_decimate_ours}
\begin{algorithmic}[1]
\renewcommand{\algorithmicrequire}{\textbf{Input:}}
\renewcommand{\algorithmicensure}{\textbf{Output:}}
\REQUIRE A triangular mesh $\mathcal{T}^i=(\mathcal{V}^i,\mathcal{F}^i)$,  the number of output vertices $N$.
\ENSURE A decimated mesh ${\mathcal{T}^o}=(\mathcal{V}^o,\mathcal{F}^o)$; vertex clustering information $\textit{replace}$; vertex degeneration information \textit{mapping}.
\STATE  {select any pair $({\bf v}_i$, ${\bf v}_j)$ that is an edge.} 
\STATE {compute the quadric of each vertex pair}
\STATE sort all pairs ascendingly based on their quadrics
\STATE set the number of vertices to remove as $N_r=|\mathcal{V}^i|-N$; the number of vertices removed as $n_r=0$
\FOR{each pair (${\bf v}_i$, ${\bf v}_j$)}
\IF{${\bf v}_i$, ${\bf v}_j$ are not in any cluster \AND $n_r<N_r$}
\STATE (a) form $({\bf v}_i, {\bf v}_j)$ as a new cluster
\STATE (b) set $n_r=n_r+1$
\ENDIF
\ENDFOR
\STATE add the remaining vertices that are not in any cluster to an existing cluster containing vertices connected to it.
\FOR{each vertex cluster}
\STATE (a) compute quadric $Q$
\STATE (b) determine the target position $\bar{\bf v}$ of contraction
\STATE (c) perform contraction $({\bf v}_i, {\bf v}_j) \rightarrow \bar{\bf v}$
\ENDFOR
\STATE return
 \end{algorithmic}
 \end{algorithm}
 \vspace{-1mm}
\section{Mesh Convolution}
\vspace{-1mm}
Let $\mathcal{T}=(\mathcal{V},\mathcal{F})$ be a triangular mesh.
The input features of its vertices are $(x,y,z,r,g,b)$, while the normals and areas of each facet are ${\bf n}=(n_x,n_y,n_z)$ and $A$.
If the mesh is rendered, we denote the textures of a facet as $\Gamma\times3$, in which $\Gamma$ represents the texture resolution of the facet, $3$ relates to the $(r,g,b)$ colors. Since the facets usually have varying areas, we allow varying $\Gamma$ for different facets as well. 

\vspace{-1mm}
\subsection{Vertex2Facet convolution} 
\vspace{-1mm}
We compute the features of each facet based on the features of its vertices. In particular, we define a kernel composed of 3 filters associated to the three vertices of the triangular facet. The Barycentric interpolation is used to incorporate fuzzy scheme into the convolution. We determine the total number of interpolated points $K$ as 
\begin{equation}\label{equ:interpolate_num}
K=\frac{k\times(k+1)}{2},~~ 
k = \left\lfloor \frac{A-A_{\min}}{A_{\max}-A_{\min}}  \right\rfloor \times \alpha + \beta,\\    
\end{equation}
in which $A_{\min}, A_{\max}$ are the minimum and maximum facet areas of the mesh, while $\alpha, \beta$ are the hyper-parameters. We use 
$\alpha=\beta=1$ in our experiments.

Let $(I_1,I_2,I_3)$ be the vertex features, and $(w_1,w_2,w_3)$ be the filter weights. The Barycentric coordinates satisfies $\xi_{k1}+\xi_{k2}+\xi_{k3}=1,~\xi_{k1},\xi_{k2},\xi_{k3}\geq0$. We use it to interpolate interior points uniformly on the facet, whose features and filter weights are computed respectively as 
\begin{align}
I_k&=\xi_{k1}*I_1+\xi_{k2}*I_2+\xi_{k3}*I_3,\\
W_k&=\xi_{k1}*w_1+\xi_{k2}*w_2+\xi_{k3}*w_3, ~k\in K.
\end{align}
Here, $K$ is the total number of interpolated points defined based on the facet area. 
With $\{I_k\}$ and $\{W_k\}$, 
we finally compute the feature of each facet as
\begin{align}
\vspace{-5mm}
J&=\frac{1}{K}\sum_k {W_k*I_k}.
\vspace{-5mm}
\label{eq:V2F_Conv}
\end{align}
\noindent\textbf{Facet2Facet convolution:}
The facet2facet convolution is only applicable when the input mesh is rendered with textures. 
Each facet contains $\Gamma$ interpolated points with texture features of size $\Gamma\times3$. 
The definition of the facet2facet convolution is quite similar to that of the vertex2facet convolution. The main difference is that in facet2facet convolution, there is no need to interpolate features for the interior points since their features are already available. We only need to interpolate the filter weights $\{W_k\}$, and then compute the facet features following 
Eq.~(\ref{eq:V2F_Conv}). 



\begin{figure}[!t]
    \centering
    \begin{tabular}{cccc}
    \hspace{-2mm}
   \includegraphics[width=0.1\textwidth]{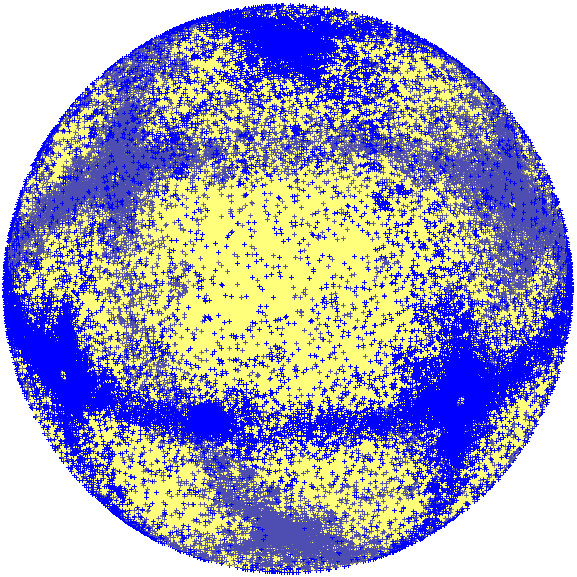} &\hspace{-2mm} \includegraphics[width=0.1\textwidth]{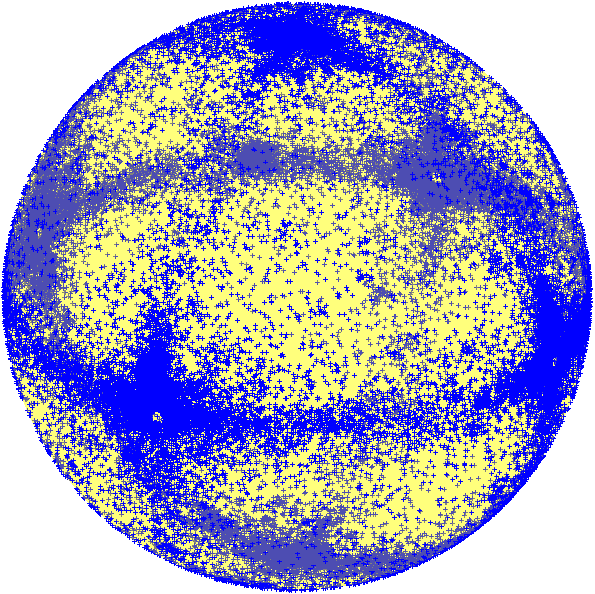} &\hspace{-2mm}
  \includegraphics[width=0.1\textwidth]{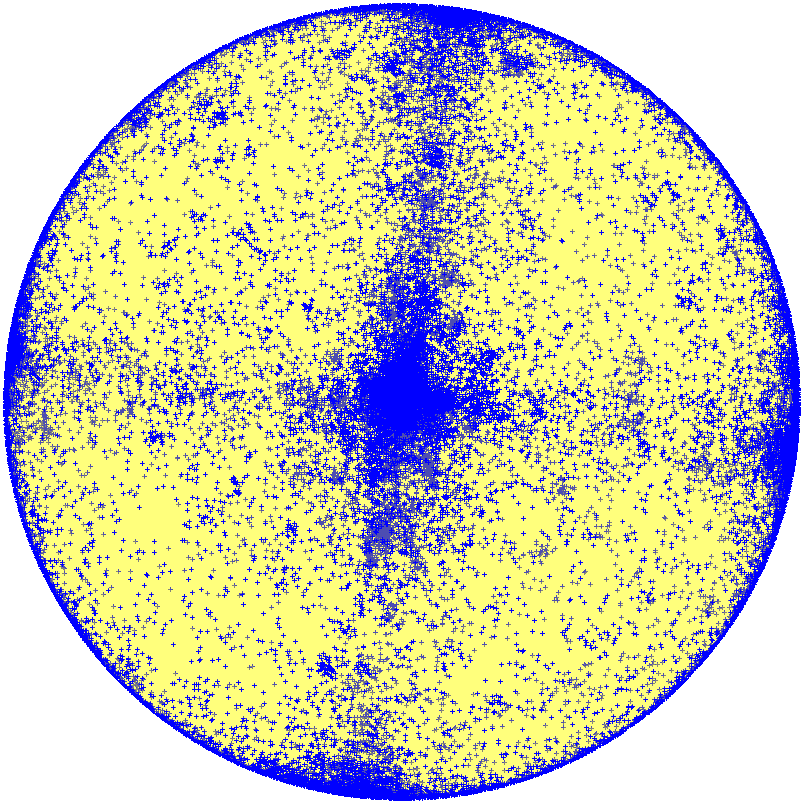} &\hspace{-2mm} \includegraphics[width=0.1\textwidth]{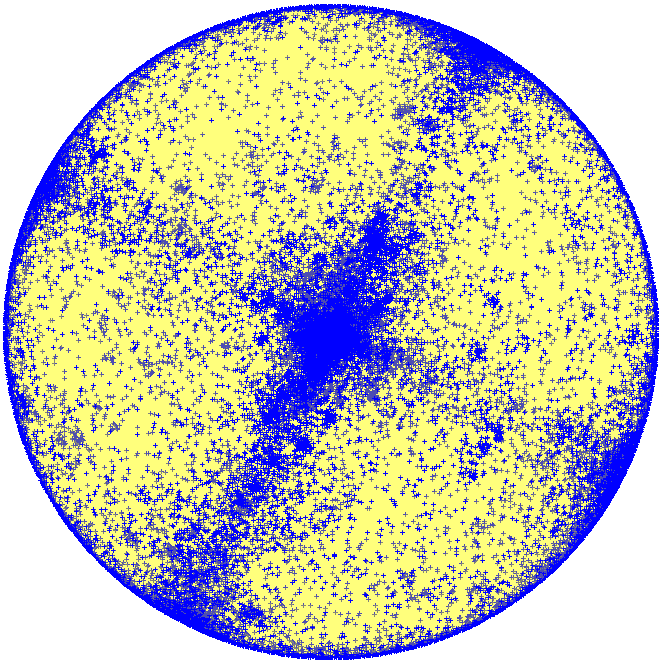} \\
  \hspace{-2mm}(a) front &\hspace{-2mm}(b) back &\hspace{-2mm}(c) top &\hspace{-2mm}(d) bottom\\
    \end{tabular}
    \vspace{-3mm}
    \caption{The distributions of surface normals of an arbitrary room. We show it from different views, including front, back, top and bottom. The six apparent clusters have centers approximating to $[\pm1,0,0],~[0,\pm1,0],~[0,0,\pm1]$.}
    \label{fig:normal_distribute}
    \vspace{-5mm}
\end{figure}
\subsection{Facet2Vertex Convolution} 
We compute vertex features from the features of their adjacent facets. 
Considering that the normals of each facet are strictly located on the surface of a unit sphere, we define filter weights of our kernel by associating them to different positions on the sphere. Besides, we observe that normals of the real-world meshes generally distribute in distinctive patterns on a unit sphere, especially for the indoor meshes. This is related to the construction preferences of human-beings. 
We show an example in Fig.~\ref{fig:normal_distribute}. The data is taken from S3DIS~\cite{armeni20163d}. There are six main clustering patterns. They correspond to different normal directions, which are roughly $[\pm1,0,0],[0,\pm1,0],[0,0,\pm1]$.
Consequently, we exploit the mixture of different Gaussian components to divide the sphere surface and implicitly cluster normals.

Let the total number of Gaussian components be $T$, their expectations and covariance matrices be $\{{\boldsymbol \mu}_t\}$ and $\{\Sigma_t\}$. Using its normal ${\bf n}_i$, we compute the fuzzy coefficients $\{\pi_{it}\}$ of the facet ${\bf f}_i$ as 
 \begin{align}
&z_{it} = ({\bf n}_i-{\boldsymbol \mu}_t)^T\Sigma_t^{-1}({\bf n}_i-{\boldsymbol \mu}_t), \\
&\pi_{it} = \frac{\exp(-z_{it})}{\sum_m \exp(-z_{im})}.
\end{align}
For simplicity, we use homogeneous diagonal matrix for $\Sigma_k$ in our experiments. Let the filter weights in the kernel be $\{w_t\}$, the adjacent facets of vertex ${\bf v}$ be $\mathcal{N}(\bf v)$, and the facet features be $\{J_{i}\}$. The vertex feature gets computed as 
\begin{align}
\vspace{-5mm}
&I_v =\frac{1}{\mathcal{N}({\bf v})}\sum_{{\bf f}_i\in\mathcal{N}({\bf v})}\Big(\sum_{t=1}^{T}\pi_{it} {w}_{t}\Big){J_{i}}. 
\vspace{-5mm}
\end{align}
In our library, we allow the expectations and covariance matrices of the Gaussians to be constant, or learnable together with the filter weights during training. We initialize the Gaussian means as regularly distributed points on the sphere in our experiments and keep those constant, while allowing the covariance matrices as learnable parameters. We note that the facet2vertex convolution is scale and translation invariant but not rotation invariant because its fuzzy coefficients are computed based on facet normals.
\begin{figure}[!t]
    \centering
    \includegraphics[width=0.46\textwidth]{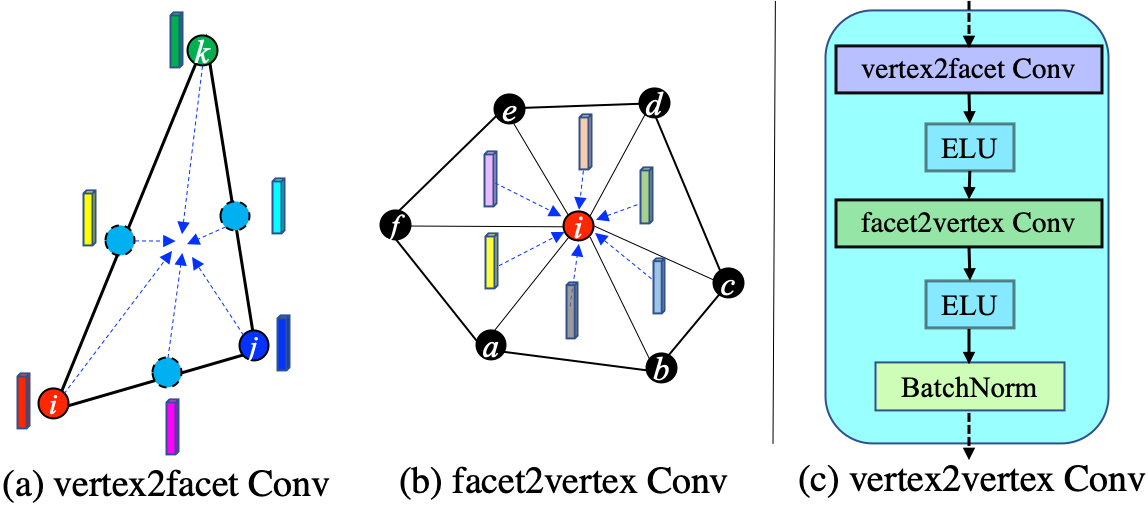}
    \caption{Convolution operations introduced in Picasso. 
    (a) The vertex2facet convolution propagates features from the original and interpolated points on a facet to the facet itself. 
    (b) The facet2vertex convolution propagates features from the adjacent facets of a vertex to the vertex itself.
    We omit illustration of the facet2facet convolution for its visual similarity to the vertex2facet convolution. (c) The vertex2vertex convolution is composed of a vertex2facet convolution followed by a facet2vertex convolution. We apply batch normalization to the vertex features.} 
    \label{fig:v2v_conv}
    \vspace{-5mm}
\end{figure}
\begin{figure*}[!t]
    \centering
    \includegraphics[width=0.88\textwidth]{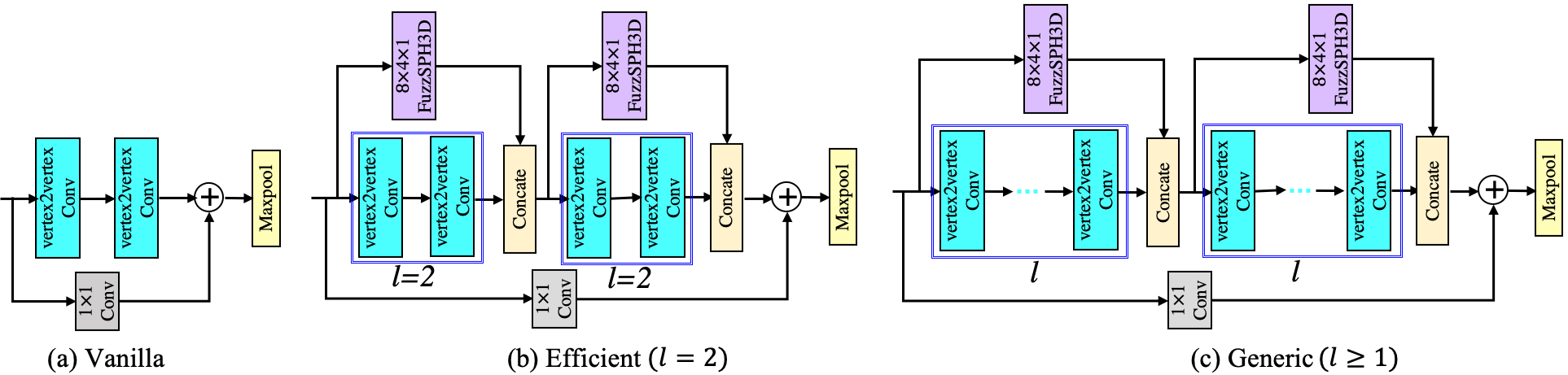}
    \caption{Convolutional blocks of PicassoNet. The network consists of 5 mesh resolutions including the input, which relates to a network of 5 hierarchical layers.  Convolutional blocks are used to learn features within a single hierarchical layer. We test the performance of different block configurations for feature learning. 
    (a) The \textit{vanilla} block consists of two  vertex2vertex convolutions with a shortcut connection.
    (b) The \textit{efficient} block type contains two vertex2vertex convolutions as well as one point-based convolution before each feature concatenation. It can process meshes with \textit{1 million} facets per-second at the inference time, while maintaining highly competitive results.
     (c) The \textit{generic} convolution block uses $l\geq 1$ vertex2vertex convolutions together with a single point-based convolution before the feature combination. We report at-par results to state-of-the-art on S3DIS~\cite{armeni20163d} by using generic block with $l=4$.} 
    \label{fig:picasso_net}
    \vspace{-5mm}
\end{figure*}

For all the proposed convolutions, we split the operation into channel-wise and depth-wise operations. Doing so is known to improve the  computational efficiency of the operation~\cite{chollet2017xception,lei2020spherical}.
Our mesh decimation technique forces the vertex number, rather than the facet number, of each decimated mesh in a batch to be the same. Therefore, we apply batch normalization to vertex features for stable computation of the batch statistics. 
Combining the vertex2facet and facet2vertex convolutions, we can perform vertex2vertex convolution. 
Figure~\ref{fig:v2v_conv} illustrate the concept of  vertex2facet and facet2vertex convolutions, along with a flow chart to elucidate the operations in vertex2vertex convolution.

\section{Network Architecture}
\vspace{-2mm}
\noindent\textbf{Receptive field:} All the convolutions we propose follow the natural organisation of vertex lists in the mesh facets. Therefore, the operations   
accumulate context from the first-order neighborhood. This is similar to the receptive field reached by $3\times3$ convolution in standard CNNs for image processing. To benefit feature learning from larger context, we apply multiple cascaded vertex2vertex convolutions in our network, which leads to a deeper neural network. 

\noindent\textbf{Disconnected components:} Real-world meshes are not guaranteed to form a connected graph. It is probable that a mesh is composed of multiple disconnected components. This breaks the context aggregation between different components for mesh convolutions.
On the other hand, \textit{forcing} the mesh to be connected can damage its geometric structure. In this work, we counter the potential issues of mesh connections using point cloud convolution, e.g.~the fuzzy SPH3D~\cite{lei2020seggcn}.
By using range search \cite{preparata2012computational}, point cloud convolutions permit arbitrary respective fields. Finally, this strategy enables our feature learning to go beyond single connected components  and learn features across multiple components. 

\noindent\textbf{The PicassoNet:} To explore the proposed modules, we design a U-net \cite{ronneberger2015u} like architecture for the popular semantic segmentation task. For consistency, we name our network as PicassoNet. The proposed network is composed of 5 mesh resolutions, indicating 5 hierarchical layers. We use convolution blocks to learn features within a single network resolution. 
Figure~\ref{fig:picasso_net} shows different block configurations considered in this work. The \textit{vanilla} block uses only 
the mesh convolutions, while the \textit{efficient} block uses $l=2$ mesh convolutions accompanied by a point-based convolution in parallel, before feature concatenation.  PicassoNet based on the efficient block is able to process 1 million facets per-second at the inference time, still reporting  competitive results.
The \textit{generic} block uses arbitrary number of mesh-based convolutions with one point-based convolution to extract features before the  concatenation. Unlike DCM-Net \cite{schult2020dualconvmesh}, we do not tune the ratio of feature channels between mesh-based and point-based convolutions. We use identical feature channels for both the convolutions, which correspond to a fixed ratio of 0.5. We use max mesh pooling to obtain down-sampled features for the low-resolution networks. For efficiency, we apply the convolution blocks only in the encoder. Whereas in the decoder, we use  $1\times1$ convolutions and mesh unpooling to upsample the features, somewhat similar to the methods in  \cite{lei2020seggcn,thomas2019KPConv}.
\vspace{-2mm}
\section{Experiments}
\vspace{-1mm}
PicassoNet takes 6-dimensional  input features $(x,y,z,r,g,b)$, similar to many point cloud networks \cite{lei2020seggcn,lei2020spherical,thomas2019KPConv}. Although normals are readily available for both facets and vertices on meshes, we do not observe noticeable performance gain by using vertex normals as input features in our experiments. To investigate the modules presented in the Picasso library, we focus on the semantic segmentation of the real-world dataset S3DIS~\cite{armeni20163d} using the PicassoNet. We choose S3DIS dataset due to the public availability of its training and testing sets. The $rgb$ color values are re-scaled in the range $[-1,1]$ before feeding into the network.

\begin{figure}
    \centering
    \includegraphics[width=0.46\textwidth]{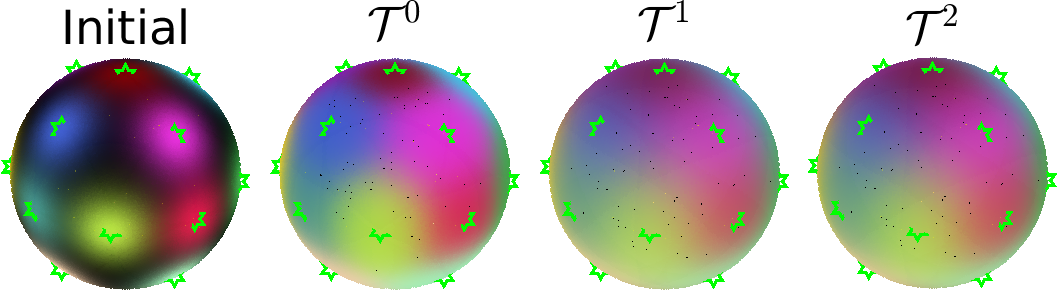}
    \vspace{-2mm}
    \caption{Visualized comparison of the trained Gaussians at different mesh resolutions with their universal initialisation. The brightness of the colors also accounts for the difference in the values. The green markers indicates centers of each Gaussian. We show only the results at mesh resolutions $\mathcal{T}^0,~\mathcal{T}^1,~\mathcal{T}^2$. The following resolutions look visually similar to $\mathcal{T}^2$.}
    \label{fig:trained_gmms}
    \vspace{-7mm}
\end{figure}

Our network is trained on a single GeForce RTX 2080 Ti GPU with Adam Optimizer \cite{kingma2015adam}. For training, we adopt initial learning rate 0.001 with exponential decay. In specific, we decay the learning rate with a rate of 0.7 every 20K batch updates.  
Throughout the experiments, we use 18 Gaussian components whose centers are located regularly on the unit sphere for all facet2vertex convolutions.
See supplementary for their specific values. The standard deviations of these Gaussians are universally initialized to 0.25, which is determined by the nearest center distances between different Gaussians. 
We switch off training of the Gaussian expectations on purpose because we observe that their values do not change too much in our experiments. Figure~\ref{fig:trained_gmms} provides an example to show the updates of the standard deviations of the Gaussians during the training. 

We use the fuzzy spherical kernel proposed in \cite{lei2020seggcn} to achieve point-based convolution in our \textit{efficient} and \textit{generic} blocks. The kernel size and neighborhood search configurations are identical to \cite{lei2020seggcn}. 
Our network uses batch size $16$ in the training and takes point clouds of size $8,192$ as inputs.
We exploit the widely used data augmentations in our experiments, including random flipping, shifting, random scaling, noisy translation, random azimuth rotation and arbitrary rotation perturbations. We also apply random color dropping to augment the vertex textures. The shuffling strategy we inserted in the mesh simplification also played a role of data augmentation. We apply these augmentations on-the-fly during the network training. 

\vspace{1mm}
\noindent{\textbf{Network Configuration:}}
We use identical feature configurations for different convolution blocks shown in Fig.~\ref{fig:picasso_net}. In specific, the desired vertex sizes of meshes $\mathcal{T}^0$, $\mathcal{T}^1$, $\mathcal{T}^2$, $\mathcal{T}^3$, $\mathcal{T}^4$ are $8192$, $2048$, $1024$, $512$, $256$ respectively. 
To construct the neighborhoods for point-based convolution, we use an increasing range search radius $0.1$, $0.2$, $0.4$, $0.8$, $1.6$.  
The output feature sizes of the 5 hierarchical layers are 128, 128, 256, 256, 256, respectively. When mesh-based and point-based convolutions are both included, they each compute features of half the size of the defined output feature channels. For example, when the output feature channels are set to 128, they each compute features of 64 channels. This results in their concatenated features to have 128 channels. We set the multiplier $\lambda$ of all convolutions to 1 except for the convolution performed on the inputs, which has $\lambda=2$. 
In the decoder, we explore mesh unpooling for feature interpolation. The output feature channels of the $1\times 1$ convolutions are the same as their corresponding encoder features. 
PicassoNet also classifies the features obtained at resolution $\mathcal{T}^0$ in the decoder directly for efficiency.

\begin{table*}[!t]
\caption{Performance of different configurations for PicassoNet on the fifth fold (Area 5) of S3DIS dataset. The results of PicassoNet using $l=4$ outperforms SegGCN and DCM-Net slightly, and are  competitive for KPConv. See supplementary for details of input representations exploited by each method.}\label{tab:s3dis_seg}
\vspace{-2mm}
\begin{adjustbox}{width=1\textwidth}
{\Huge\begin{tabular}{c|l|ccc|ccccccccccccc}
  \hline
 &Methods& OA& mAcc & mIoU & ceiling & floor & wall & beam & column & window & door & table & chair & sofa & bookcase & board & clutter \\
  \hline
\multirow{11}{*}{\rotatebox[origin=c]{90}{Area 5}}
& PointNet \cite{qi2017pointnet}& - &49.0 &41.1 &88.8 &97.3 &69.8 &0.1 &3.9 &46.3 &10.8 &58.9 &52.6  &5.9 &40.3  &26.4 &33.2\\
&SEGCloud \cite{tchapmi2017segcloud} & - &57.4 &48.9 &90.1 &96.1 &69.9 &0.0 &18.4 &38.4 &23.1 &70.4 &75.9 &40.9 &58.4 &13.0 &41.6\\
&Tangent-Conv \cite{tatarchenko2018tangent}& 82.5 &62.2 &52.8 &- &- &- &- &- &- &- &- &- &- &- &- &-\\
&SPG \cite{landrieu2017large} & 86.4 &66.5 &58.0 &89.4 &96.9 &78.1 &0.0 &42.8 &48.9 &61.6&75.4 &84.7 &52.6 &69.8  &2.1 &52.2\\
&PointCNN \cite{li2018pointcnn}& 85.9& 63.9& 57.3& 92.3& 98.2 &79.4& 0.0 &17.6& 22.8& 62.1& 74.4& 80.6& 31.7& 66.7& 62.1& 56.7\\
&SSP+SPG \cite{landrieu2019point}& 87.9 & 68.2 &61.7&- &- &- &- &- &- &- &- &- &- &- &- &-\\
&GACNet \cite{wang2019attention}& 87.8 & - &62.9&92.3 &98.3 &81.9 &0.0 &20.4 &59.1 &40.9 &78.5 &85.8 &61.7 &70.8 &74.7 &52.8\\
&SPH3D-GCN \cite{lei2020spherical}& 87.7 &65.9 &59.5 &93.3 &97.1 &81.1 &0.0 &33.2 &45.8 &43.8  &79.7 &86.9 &33.2 &71.5  &54.1 &53.7\\
&KPConv \cite{thomas2019KPConv}& - &70.9 &65.4 &92.6 &97.3 &81.4 &0.0 &16.5 &54.5 &69.5 &90.1 &80.2 &74.6 &66.4  &63.7 &58.1\\ 
&SegGCN \cite{lei2020seggcn} & 88.2 &70.4 &63.6 &93.7 &98.6 &80.6 &0.0
&28.5 &42.6 &74.5  &80.9
&88.7
&69.0
&71.3  &44.4 &54.3\\
&DCM-Net\cite{schult2020dualconvmesh} & - & 71.2 & 64.0 &92.1 &96.8 &78.6 &0.0 &21.6 &61.7 &54.6 &78.9 &88.7 &68.1 &72.3 &66.5 &52.4\\
\hline
&PicassoNet (Prop. vanilla) & 86.6 &65.6 &58.0 &93.2 &98.0 &78.3 &0.0
&16.8 &28.5 &61.7  &75.5
&86.4
&49.2 &69.0&45.4&52.1\\
&PicassoNet (Prop. $l=1$) & 87.8 &67.8 &61.0 &93.8 & 97.6 & 80.6 & 0.0 & 32.6 & 48.2& 69.3 & 77.8& 87.0& 56.8 & 70.1 & 25.5 & 53.3\\
&PicassoNet (Prop. $l=2$) & 88.2 &69.4 &62.5 &93.2 &98.4 &81.1 &0.0
&32.1 &45.9 &75.6  &78.5
&85.9
&51.5  &69.0&45.2&55.3\\
&PicassoNet (Prop. $l=4$) & 89.4 &70.9 &64.6 &93.3 &97.7 &83.5 &0.0
&31.9 &53.4 &69.2  &81.7
&88.0
&50.5
&74.3  &58.2 &57.9\\
\hline
\end{tabular}}
\end{adjustbox}
\end{table*}
\begin{table*}[!thb]
    \centering
    \caption{Training and testing time of the efficient PicassoNet ($l=2$) for different mini-batch settings. The `mesh size' indicates total number of vertices and facets as (vertices, facets) in the concatenated mesh of a batch. We add the testing runtime of SegGCN under identical settings for pure point cloud input at the bottom row as a reference.}\label{tab:PicassoNet_runtime}
    \vspace{-1mm}
    \begin{adjustbox}{width=1\textwidth}
    {\begin{tabular}{l|c|c|c|c|c|c}
    \hline
  Phase &  training & \multicolumn{5}{c}{testing} \\
  \hline\hline
 batch size & 16& 16& 32& 64 & 68& 128 \\
 \hline
 mesh size & ($0.13$M,$0.24$M)& ($0.13$M,$0.24$M)&($0.26$M,$0.48$M) &($0.52$M,$0.96$M) & ($0.56$M,${\bf 1.02}$M)& ($1.04$M,$1.92$M) \\
 \hline
  runtime & $920$ ms& $318$ ms& $500$ ms & $890$ ms& ${\bf 915}$ ms&$1750$ ms\\
        \hline
        \hline
   \multicolumn{2}{c|}{testing runtime of SegGCN~\cite{lei2020seggcn}}& 153 ms & 250 ms & 470 ms & 485 ms & 930 ms \\
        \hline
    \end{tabular}}
    \end{adjustbox}
    \vspace{-5mm}
\end{table*}

\subsection{Semantic Segmentation}
\vspace{-1mm}
The Stanford large-scale 3D Indoor Spaces (S3DIS) dataset~\cite{armeni20163d} is a real-world dataset composed of dense 3D point clouds but sparse 3D meshes of 6 large-scale indoor areas. The data is collected using Matterport scanner from three different buildings on the Standard campus. Semantic segmentation on this dataset is to classify the 13 defined classes, \emph{ceiling, floor, wall, beam, column, window, door, table,
chair, sofa, bookcase, board}, and \emph{clutter}. We follow the standard training/test split by using Area 5 as the test set while the other areas as training set \cite{landrieu2017large,li2018pointcnn,qi2017pointnet,tchapmi2017segcloud,wang2019graph}.
The evaluation metrics for this dataset comprise the Overall Accuracy (OA), average Accuracy of all 13 classes (mAcc),  class-wise Intersection Over Union (IoU), together with their average (i.e.~mIoU). mIoU is commonly considered a more reliable metric than the others. 

DCM-Net \cite{schult2020dualconvmesh} used over-tessellation and interpolation to produce high resolution meshes with labels from the original meshes in S3DIS. In contrast, we triangulate the labelled point cloud into triangular meshes using the algorithm from  \cite{pointcloud2mesh}. 
Our network takes meshes of size 
$2.0m\times2.0m$ cropped from the room mesh as input data.
The experiment results for different block configurations in the PicassoNet are provided in Table~\ref{tab:s3dis_seg}.
It can be noticed that PicassoNet ($l=4$) provides competitive results to KPConv \cite{thomas2019KPConv}, and slightly outperforms SegGCN \cite{lei2020seggcn} and DCM-Net \cite{schult2020dualconvmesh}.
Table~\ref{tab:PicassoNet_runtime} reports the training time of PicassoNet ($l=2$) for batch size 16, and its testing time for different batches. It can be seen that PicassoNet ($l=2$) can process over a million facets per-second on a commonly available single RTX 2080 Ti GPU. 
\subsection{Mesh Simplification Efficiency}
\begin{figure}[!t]
\centering
\begin{tabular}{cc}
\includegraphics[width=0.47\textwidth]{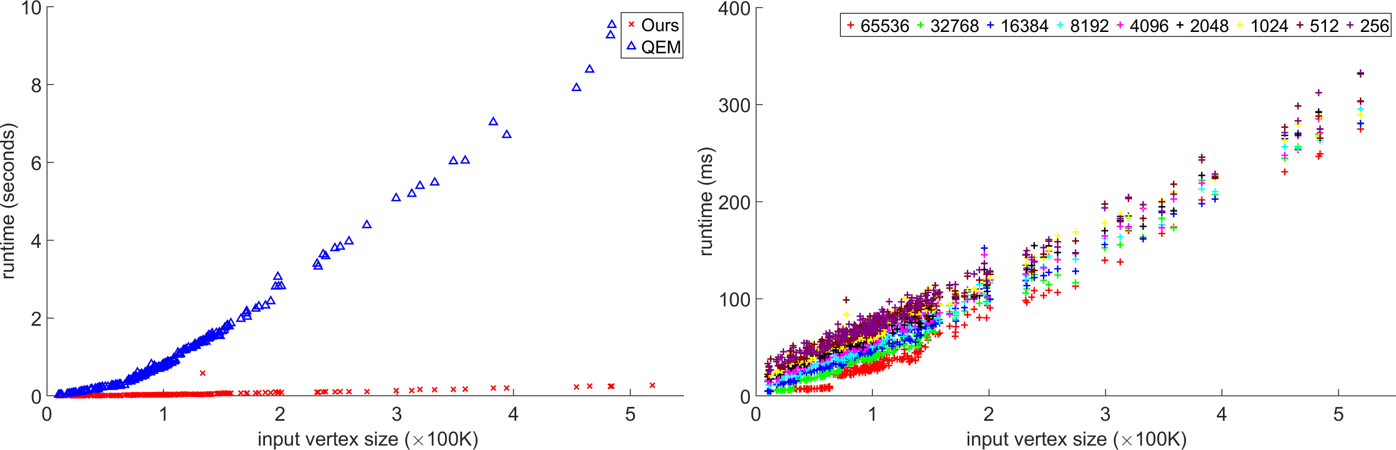}
\end{tabular}
\caption{The left figure compares the runtime of our mesh decimation method with QEM. The data is plotted by decimating every sample in S3DIS to a mesh of 65536 vertices by both methods. We notice that QEM takes much more time than ours, usually in `seconds', while  our algorithm runs in `milliseconds'. The right figure shows the time taken by our simplification algorithm to decimate all mesh samples in S3DIS to different vertex sizes, including 65536, 32768, 16384, 8192, 4096, 2048, 1024, 512, and 256. It can be observed that the runtime of our algorithm remains  generally linear to the vertex size of the input mesh.}
\label{fig:QEM_ours_timeCompare}
\vspace{-6mm}
\end{figure}

We summarize the statistics of vertex and facet sizes of room meshes we generate in S3DIS. The minimum, maximum, average of (vertex, facet)
numbers are $(10K,20K)$, $(1157K,2288K)$, and $(125K,247K)$. The standard deviations are as large as $(122K,241K)$. 
Our fast mesh decimation method contributes largely to the efficiency of our network, e.g.~million facets processed per second. We apply our simplification algorithm to all room samples in S3DIS to decimate them each into a mesh of 65536 vertices. We compare the speed of our algorithm to the quadric error metrics (QEM) method~\cite{garland1997surface} that our method builds upon. The results are shown in Fig.~\ref{fig:QEM_ours_timeCompare} (left plot), in which the $x$-axis indicates input vertex sizes of the room meshes while the $y$-axis reports the runtime of the algorithms.

Besides, we also test the runtime of our algorithm by decimating those room meshes into different resolutions, which corresponds to vertex sizes of 
65536, 32768, 16384, 8192, 4096, 2048, 1024, 512 and 256. These results are reported in Fig.~\ref{fig:QEM_ours_timeCompare} (right plot). 
We can tell from the figure that the runtime of our algorithm is essentially linear in the size of vertices of the input mesh. 
These conclusions about the efficiency of our simplification algorithm are generic, as we observe similar results consistently on other mesh datasets also. See the supplementary for further results. 


\section{Conclusion}
We provide a fast CUDA-accelerated mesh decimation technique to facilitate the 3D community to explore hierarchical neural architectures on meshes. We propose 3 novel convolutions, namely;  vertex2facet, facet2facet, and facet2vertex. We  provide CUDA implementations for forward and backward passes for these convolutions. We also introduce the vertex2vertex convolution on top of vertex2facet and facet2vertex convolutions. We use Gaussian mixtures and Barycentric interpolation to incorporate fuzziness into the proposed convolutions.
Our mesh simplification gather vertex clustering information into a 1-D tensor, which is convenient for the CUDA-based pooling and unpooling operations. Most importantly, we introduce Picasso, a self-contained library for deep learning over 3D meshes and  point clouds. 
We also present PicassoNet for semantic segmentation, and test its performance on S3DIS which provides at par results to state-of-the-art. The efficient version of PicassoNet can process 1 \textit{million} facets during inference while retaining competitive accuracy. 
We also observe that more mesh convolutions in the deeper version  increases the receptive field, and hence learn better features.
In the future, we will maintain and upgrade the introduced library for efficiency and further functionalities.

\noindent{\bf Acknowledgements: }
This work is supported by 
ARC Discovery Grant DP190102443.

{\small
\bibliographystyle{ieee_fullname}
\bibliography{egbib}
}
\newpage
\onecolumn
\begin{center}
\Large
\textbf{Picasso: A CUDA-based Library for Deep Learning over 3D Meshes\\\textit{Supplementary Material}}    
\end{center}
\vspace{3mm}
\section{Picasso Library}
\begin{figure*}[!h]
    \centering
    \includegraphics[width=0.72\textwidth]{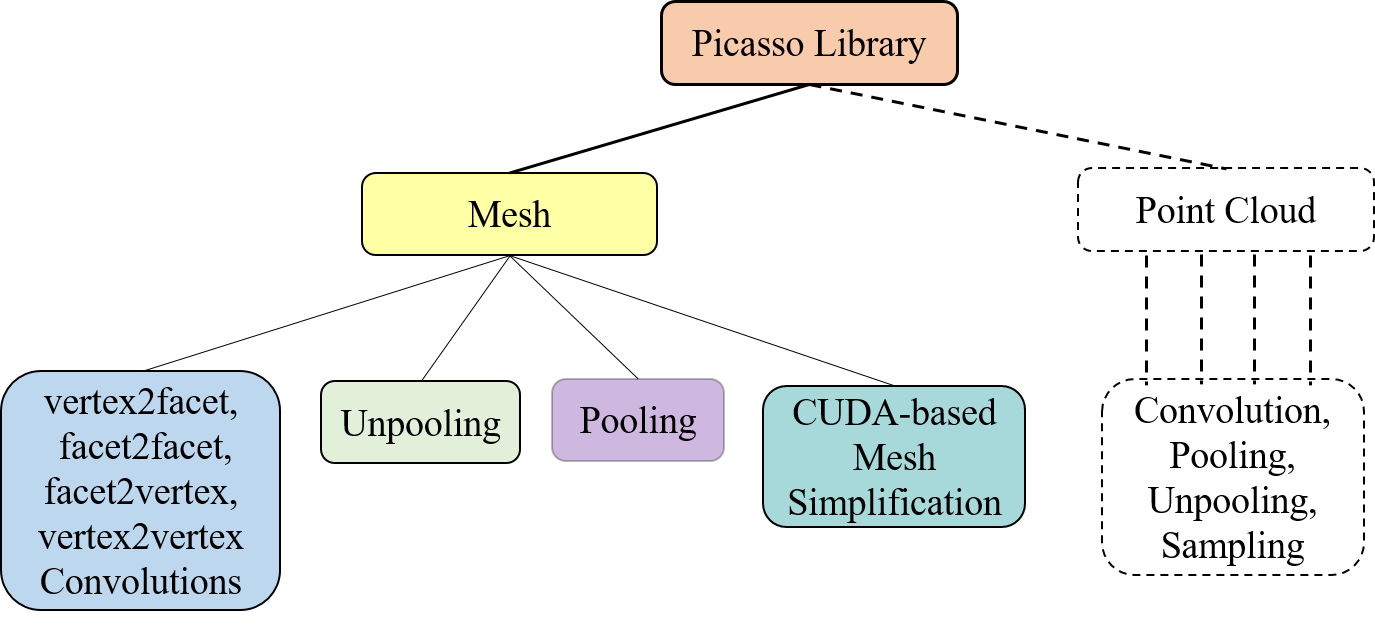}
\caption{Deep learning modules included in the Picasso Library.}
\label{fig:library_tree}
\end{figure*}

We summarize the major modules included in Picasso Library as Fig. \ref{fig:library_tree}. All the novel operations introduced in this paper are colorized, while the previous point cloud based operations [{\color{green}32, 33}] are left as blank. We also compare performance of the included convolutions for a very large input size, i.e. $65536$ in Table~\ref{tab:templates_summary}. 
For mesh-based convolutions, we use a batch of 16 meshes as input, whose vertex size each is 65536. For point based convolutions, we use only vertices of those meshes as input.
Currently, 3D deep networks are widely taking data samples of $<$10000 points/vertices as input. For spatial graph convolutions, the graph construction based on neighborhood search takes significant amount of time, especially when the input size is large. For instance, see Table 10 of  SPH3D-GCN~[{\color{green}33}]. 
We introduce mesh-based modules to take advantage of its geodesic connections and save graph construction time.
It is worth noting that we estimate the runtime under Tensorflow~2. 
\begin{table*}[!thb]
    \centering
    \caption{Runtime comparison of different mesh-based convolutions and point-based convolutions. $T$ represents the total number of filters in the kernel. $C_{in}$ denotes the number of input feature channels, and $\lambda$ is the multiplier of separable convolutions. Regardless of graph construction, the speed of mesh-based convolutions are similar to point-based convolutions. However, since
    point-based convolutions demand the graph to be pre-constructed before convolution, which takes significant amount of time when input size is large. For time concern, mesh-based convolutions are  preferred as the input size grows.  }
    \label{tab:templates_summary}
    \begin{tabular}{l|c|c|c|c|c|c}
    \hline
        \multirow{ 2}{*}{Convolution Type}& \multirow{ 2}{*}{batch size} & \multirow{ 2}{*}{input vetex/point size}& \multicolumn{3}{c|}{kernel configuration}&\multirow{ 2}{*}{runtime (ms)}\\
        \cline{4-6}
        &&&$T$&$C_{in}$&$\lambda$&\\
        \hline
        facet2facet & 16&$65536$& 3& 6 & 8&125\\
        vertex2facet & 16&$65536$& 3 & 6 & 8&89\\
        facet2vertex & 16&$65536\xrightarrow []{\text{strided}}16384$& 12 & 48 & 2 &195\\
        \hline
        hard SPH3D~[{\color{green}33}]  & 16&$65536\xrightarrow []{\text{strided}}16384$& $8\times2\times1+1$ & 48 & 2 &118~+ graph construction\\
        fuzzy SPH3D~[{\color{green}32}]  & 16&$65536\xrightarrow []{\text{strided}}16384$& $8\times2\times1+1$ & 48 & 2 &188~+ graph construction\\
        \hline
    \end{tabular}
\end{table*}
\section{Quadric error computation}
The plane function of an arbitrary facet can be denoted as ${\bf n}^\intercal {\bf x}+d=0$, where ${\bf x}=[x,y,z]^\intercal$ is the point in 3D space, ${\bf n}=[n_x,n_y,n_z]^\intercal$ is the facet normal, and $d$ is the intercept.
The quadric $Q$ of each facet is defined as $Q = ({\bf A},{\bf b},c) = ({\bf n} {\bf n}^\intercal,d{\bf n},d^2)$. 
It associates a value named \textit{quadric error} to an arbitrary point ${\bf x}$ in space , which is computed as 
\begin{align}
Q({\bf x}) = {\bf x}^\intercal {\bf A}{\bf x}+2{\bf b}^\intercal{\bf x}+c.
\end{align}
The quadric of each vertex ${\bf v}$ is an accumulation of the quadric of its adjacent facets $\mathcal{N}({\bf v})$, represented as
\begin{align}
Q_v = ({\bf A}_v, {\bf b}_v, c_v)=\left(\sum_{{\bf f}_i\in\mathcal{N}({\bf v})} {\hspace{-2mm}{\bf A}_i},~\sum_{{\bf f}_i\in\mathcal{N}({\bf v})}{\hspace{-2mm}{\bf b}_i^\intercal},~\sum_{{\bf f}_i\in\mathcal{N}({\bf v})} {\hspace{-2mm}c_i}\right).  
\end{align}
The quadric of each vertex cluster $\mathcal{C}$ to be contracted is an accumulation of the quadric of all the vertices in it, that is
\begin{align}
Q_\mathcal{C} = ({\bf A}_{\mathcal{C}}, {\bf b}_{\mathcal{C}}, c_{\mathcal{C}})=\left(\sum_{{\bf v}\in\mathcal{C}}{\bf A}_v,~\sum_{{\bf v}\in\mathcal{C}}{\bf b}_v,~\sum_{{\bf v}\in\mathcal{C}}c_v\right).    
\end{align}
The \textit{optimal vertex placement} of each cluster after contraction is ideally computed as 
\begin{align}
\bar{{\bf v}} = -{\bf A}^{-1}_\mathcal{C}{\bf b}_\mathcal{C},
\label{eq:inverse_optimal}
\end{align}
for which we determine if ${\bf A}_\mathcal{C}$ is a full-rank matrix by checking the reciprocal of its condition number. However, we still observe numerical instability in the decimated mesh. We therefore replace the computation of $\bar{{\bf v}}$ to be 
\begin{align}
\bar{{\bf v}} = \frac{1}{|\mathcal{C}|}\sum_{{\bf v}\in\mathcal{C}} {\bf v}.
\label{eq:average_optimal}
\end{align}
Open3D~[{\color{green} 71}] computes $\bar{{\bf v}}$
in the same way. 
This improves the final results noticeably. See Table~\ref{tab:s3dis_area5_bugfixed} for the performance comparison of PicassoNet ($l=2$) on S3DIS Area 5 using
Eq. (\ref{eq:inverse_optimal}) and Eq. (\ref{eq:average_optimal}).
\begin{table*}[!h]
\caption{Performance of PicassoNet ($l=2)$ on the fifth fold (Area 5) of S3DIS dataset under different computations of the optimal vertex placement $\bar{\bf v}$.  }\label{tab:s3dis_area5_bugfixed}
\begin{adjustbox}{width=1\textwidth}
{\Huge\begin{tabular}{l|ccc|ccccccccccccc}
  \hline
 Methods& OA& mAcc & mIoU & ceiling & floor & wall & beam & column & window & door & table & chair & sofa & bookcase & board & clutter \\
  \hline
PicassoNet ($l=2$, \ref{eq:inverse_optimal}) & 88.2 &69.4 &62.5 &93.2 &98.4 &81.1 &0.0
&32.1 &45.9 &75.6  &78.5
&85.9
&51.5  &69.0&45.2&55.3\\
PicassoNet ($l=2$, \ref{eq:average_optimal}) & 88.7 &69.5 &63.1 &94.8 &98.4 &81.4 &0.0 &30.4 &53.7 &71.2  &76.8 &87.5
&48.1 &69.9 &51.2&57.2\\
\hline
\end{tabular}}
\end{adjustbox}
\end{table*}
\section{Backward propagations of different convolutions}
In the main paper, we present forward computations of the vertex2facet and facet2vertex convolutions in Eqs.~({\color{red}1}),~({\color{red}2}),~({\color{red}3}) and Eqs.~({\color{red}4}),~({\color{red}5}),~({\color{red}6}), respectively. In this section, we analyze their backward propagations in Eq.~(\ref{equ:vertex2facet_analyze}) and (\ref{equ:facet2vertex_analyze}) correspondingly. 
The computations of facet2facet convolution are similar to those of vertex2facet convolution. We hence omit them here to avoid redundancy. We also provide the relate forward computations in Eq. (\ref{equ:vertex2facet_analyze}) and \ref{equ:facet2vertex_analyze}) as references. Please refer to the main paper for the notations.
Finally, as the GMM-based fuzzy coefficients $\{\pi_{it}\}$ are readily computed with built-in Tensorflow functions, there is no need for us to analyze their backward propagations manually because Tensorflow is able to achieve it automatically. 
{\large
\begin{align}
\text{vertex2facet~ kernel:}&
\begin{cases}
J=\frac{1}{K}\sum_k {W_k I_k}. \\[3mm]
\frac{\partial J}{\partial I_s}=\frac{1}{K}\sum_{k}(\xi_{ks} W_k), s\in\{1,2,3\}. \\
\frac{\partial J}{\partial w_s}=\frac{1}{K}\sum_{k}(\xi_{ks} I_k),~ s\in\{1,2,3\}.
\end{cases}
\label{equ:vertex2facet_analyze}
\\[3mm]
\text{facet2vertex kernel:}&
\begin{cases}
I_v =\frac{1}{\mathcal{N}({\bf v})}\sum_{{\bf f}_i\in\mathcal{N}({\bf v})}\Big(\sum_{t=1}^{T}\pi_{it} {w}_{t}\Big){J_{i}}. \\[3mm]
\frac{\partial I_{v}}{\partial J_i}=\frac{1}{\mathcal{N}({\bf v})}\Big(\sum_{t=1}^{T}\pi_{it} {w}_{t}\Big).\\[3mm]
\frac{\partial I_v}{\partial w_t}=\frac{1}{\mathcal{N}({\bf v})}\sum_{{\bf f}_i\in\mathcal{N}({\bf v})}(\pi_{it} J_i).\\
\end{cases}
\label{equ:facet2vertex_analyze}
\end{align}
}
\section{6-folds results on S3DIS}
We report the 6 folds results of PicassoNet ($l=2$) on S3DIS dataset in Table \ref{tab:s3dis_full}. It can be noticed that the PicassoNet using only $l=2$ layers of mesh convolutions in each block are already competitive to KPConv~[{\color{green}59}] and DCM-Net~[{\color{green}52}]. 
\begin{table*}[!h]
\caption{Performance of PicassoNet ($l=2$) on the entire 6 folds of S3DIS. PicassoNet using only $l=2$ layers of mesh convolutions in the convolution block are  competitive to KPConv and DCM-Net.}\label{tab:s3dis_full}
\begin{adjustbox}{width=1\textwidth}
{\Huge\begin{tabular}{c|l|ccc|ccccccccccccc}
  \hline
 &Methods& OA& mAcc & mIoU & ceiling & floor & wall & beam & column & window & door & table & chair & sofa & bookcase & board & clutter \\
\hline
\multirow{10}{*}{\rotatebox[origin=c]{90}{All 6 Folds}}
&PointNet [{\color{green}43}] &78.5& 66.2
&47.6 &88.0 &88.7 &69.3& 42.4& 23.1 &47.5 &51.6 &42.0 &54.1 &38.2 &9.6 &29.4 &35.2\\
&Engelmann et al. [{\color{green}1a}] &81.1 &66.4 &49.7 &90.3 &92.1 &67.9 &44.7 &24.2 &52.3 &51.2 &47.4 &58.1 &39.0 &6.9 &30.0 &41.9\\
&SPG [{\color{green}30}] & 85.5& 73.0& 62.1& 89.9
&95.1 &76.4 &62.8 &47.1 &55.3 &68.4 &73.5 &69.2 &63.2 &45.9 &8.7 &52.9\\
&PointCNN [{\color{green}35}]& 88.1& 75.6& 65.4& 94.8& 97.3& 75.8& 63.3& 51.7& 58.4& 57.2& 71.6 &69.1& 39.1& 61.2& 52.2& 58.6\\
&SSP+SPG [{\color{green}29}]& 87.9 & 78.3 &68.4&- &- &- &- &- &- &- &- &- &- &- &- &-\\
&DeepGCN [{\color{green}2a}]& 85.9& -& 60.0& 93.1& 95.3 &78.2& 33.9 &37.4& 56.1& 68.2& 64.9& 61.0& 34.6& 51.5& 51.1& 54.4\\
&KPConv~[{\color{green}59}] & - &79.1 &70.6  &93.6 &92.4 &83.1 &63.9 &54.3 &66.1 &76.6 &57.8 &64.0 &69.3 &74.9 &61.3 &60.3 \\
&SPH3D-GCN~[{\color{green}33}] & 88.6 &77.9 &68.9 &93.3 &96.2 &81.9 &58.6&55.9 &55.9 &71.7 &72.1 &82.4 &48.5 &64.5 &54.8 &60.4\\
&SegGCN~[{\color{green}32}] & 87.8 &77.1 &68.5 &92.5 &   97.6 & 78.9  &  44.6 &   58.2 &   53.7  &  67.3   & 74.6    &83.9 &   68.0 &65.7&   46.8 &   58.5\\
&DCM-Net~[{\color{green}52}] &-& 80.7&69.7& 93.7 &96.6 &81.2 &44.6 &44.9 &73.0& 73.8 &71.4 &74.3 &63.3 &63.9 &63.0 &61.9\\
\hline
&PicassoNet (Prop. $l=2$) & 89.0 &78.8 &69.8 &93.7 & 96.6 & 82.2 & 57.8 & 54.9 & 59.9 & 77.1 & 71.5 &82.5 & 51.8 &63.8 & 53.9 &61.7\\
\hline  
\end{tabular}}
\end{adjustbox}
\end{table*}

\section{Results on ScanNet}
Originally, we trained the PicassoNet using cropped meshes generated from the full-resolution meshes provided in  ScanNet dataset~[{\color{green}12}], which produced mediocre results. We found that the reason of this phenomenon is caused by high-frequency signals in noisy areas as well~[{\color{green}52}].
We hence voxelize the full-resolution meshes in ScanNet~[{\color{green}12}] using a grid size of $4cm$, and re-conducted the experiment. The performance of PicassoNet ($l=2$) on the \textit{full} and \textit{voxelized} validation set is reported in Table \ref{tab:scannet_valid}. Similar results are expected on the test set of ScanNet. 
\begin{table*}[!h]
\centering
\caption{3D semantic segmentation performance of PicassoNet ($l=2$) on the ScanNet validation set. `full' refers to results on the full-resolution meshes, while `$4cm$' refers to the results on the voxelized meshes using a voxel size of $4cm$. Similar results are expected on the test set of ScanNet. For reference, we provide the results of DCM-Net and SPH3D-GCN on the validation set, as well as the results of  popular approaches on the test set. }
\label{tab:scannet_valid}
\begin{adjustbox}{width=1\textwidth}
{\Huge\begin{tabular}{l|c|cccccccccccccccccccc}
\hline
Method & mIoU & floor &wall &chair &sofa &table& door& cab& bed &desk &toil &sink &wind& pic &bkshf &curt &show &cntr &fridg& bath &other\\
\hline
DCM-Net (VC,rad/geo)& 62.8&-&-&-&-&-&-&-&-&-&-&-&-&-&-&-&-&-&-&-&-\\
SPH3D-GCN (full)&60.0&93.9&76.6&86.4&76.7&69.0&40.1&55.6&68.4&55.3&85.3&58.8&51.0&6.8&42.2&57.2&64.8&56.2&38.9&79.5&36.1\\
SPH3D-GCN ($3cm$)&61.2&95.5&   78.3&   86.9&   78.9&   71.4&  41.6&   57.1&   70.6&   57.4&   87.0&   58.9&49.2&    7.7&   42.2&  57.8&  63.8&   58.7&  42.2&  82.1&   37.2\\
\hline
PicassoNet ($l=2$, full) & 62.5&94.4&78.0&86.5&75.3&69.4&47.4&54.1&68.8&56.6&88.4&62.6&50.6&18.2&54.4&63.4&64.8&54.1&41.6&80.4&40.3\\
PicassoNet ($l=2$, $4cm$) & 64.3&96.1&80.4&87.0&78.2&73.2&50.3&56.6&72.0&59.1&90.0&64.1&49.9&19.2&53.9&63.9&63.7&57.1&44.9&83.5&42.0\\
\hline 
\hline
ScanNet~[{\color{green}12}]&30.6&78.6&43.7&52.4&34.8&30.0&18.9&31.1&36.6&34.2&46.0&31.8&18.2&10.2&50.1&0.2&15.2&21.1&24.5&20.3&14.5\\
PointNet++~[{\color{green}44}]&33.9&67.7&52.3&36.0&34.6&23.2&26.1&25.6&47.8&27.8&54.8&36.4&25.2&11.7&45.8&24.7&14.5&25.0&21.2&58.4&18.3\\
SPLATNET$_{\text{3D}}$~[{\color{green}56}]& 39.3&92.7&69.9&65.6&51.0&38.3&19.7&31.1&51.1&32.8&59.3&27.1&26.7&0.0&60.6&40.5&24.9&24.5&0.1&47.2&22.7\\
Tangent-Conv~[{\color{green}57}]& 43.8&91.8&63.3&64.5&56.2&42.7&27.9&36.9&64.6&28.2&61.9&48.7&35.2&14.7&47.4&25.8&29.4&35.3&28.3&43.7&29.8\\
PointCNN~[{\color{green}35}] &45.8&94.4&70.9&71.5&54.5&45.6&31.9&32.1&61.1&32.8&75.5&48.4&47.5&16.4&35.6&37.6&22.9&29.9&21.6&57.7&28.5\\
PointConv~[{\color{green}65}]& 55.6&94.4&76.2&73.9&63.9&50.5&44.5&47.2&64.0&41.8&82.7&54.0&51.5&18.5&57.4&43.3&57.5&43.0&46.4&63.6&37.2\\
SPH3D-GCN~[{\color{green}33}]& 61.0&93.5&77.3&79.2&70.5&54.9&50.7&53.2&77.2&57.0&85.9&60.2&53.4&4.6&48.9&64.3&70.2&40.4&51.0&85.8&41.4\\
KPConv~[{\color{green}59}]& 68.4&93.5&81.9&81.4&78.5&61.4&59.4&64.7&75.8&60.5&88.2&69.0&63.2&18.1&78.4&77.2&80.5&47.3&58.7&84.7&45.0 \\
SegGCN~[{\color{green}32}] & 58.9&93.6&77.1&78.9&70.0&56.3&48.4&51.4&73.1&57.3&87.4&59.4&49.3&6.1&53.9&46.7&50.7&44.8&50.1&83.3&39.6\\
DCM-Net~[{\color{green}52}] & 65.8&94.1&80.3&81.3&72.7&56.8&52.4&61.9&70.2&49.4&82.6&67.5&63.7&29.8&80.6&69.3&82.1&46.8&51.0&77.8&44.9\\
 \hline
\end{tabular}}
\end{adjustbox}
\end{table*}
\section{Mesh Simplification Efficiency}
\begin{figure}[!h]
\centering
\begin{tabular}{cc}
\includegraphics[width=0.7\textwidth]{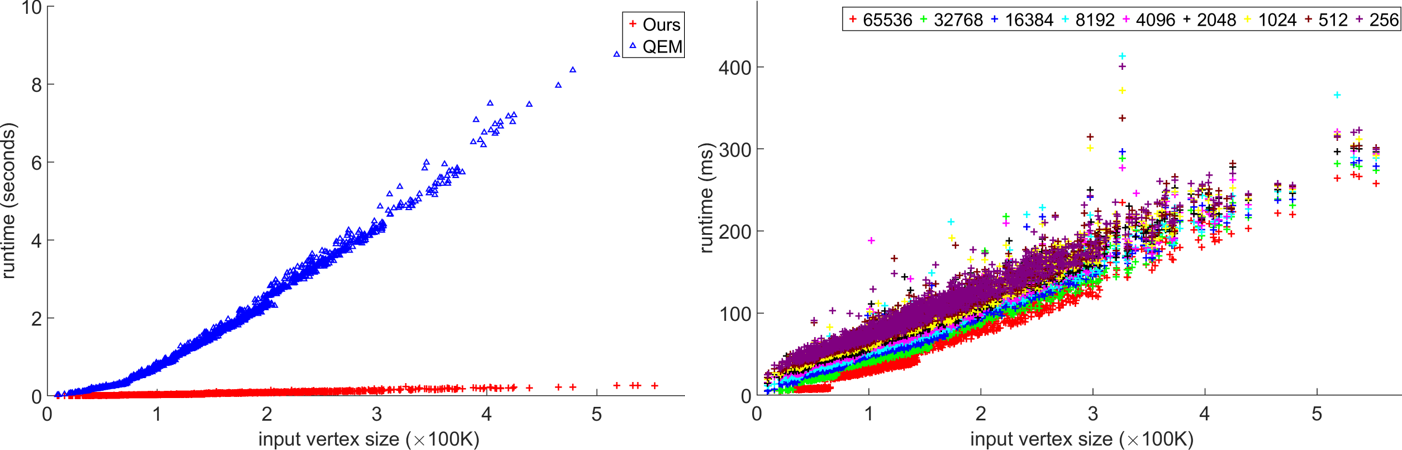}
\end{tabular}
\caption{The left figure compares the runtime of our mesh decimation method with QEM. The data is plotted by decimating every sample in ScanNet to a mesh of 65536 vertices by both methods. The right figure shows the time taken by our simplification algorithm to decimate all mesh samples in ScanNet to different vertex sizes, including 65536, 32768, 16384, 8192, 4096, 2048, 1024, 512, and 256.}
\label{fig:QEM_ours_timeCompare_scannet}
\vspace{-5mm}
\end{figure}

We summarize the statistics of vertex and facet sizes of the full-resolution meshes in ScanNet~[{\color{green}12}]. The minimum, maximum, average of (vertex, facet) number are $(9K,16K)$, $(553K,1063K)$, and $(150K,286K)$. The standard deviations are $(81K,155K)$. 
Similarly, we apply the proposed simplification algorithm to decimate all room samples in ScanNet  into a mesh of 65536 vertices. The runtime comparison of our algorithm and QEM are shown in the \textit{left} plot of Fig.~\ref{fig:QEM_ours_timeCompare_scannet}.
We also test the runtime of our algorithm while decimating the room meshes into different resolutions, for which we use identical configurations to those for S3DIS. The results are shown in the \textit{right} plot of Fig.~\ref{fig:QEM_ours_timeCompare}. 
It can be noticed from the figure that our conclusions based on S3DIS dataset consistently hold for ScanNet. 
\begin{table*}[!b]
\centering
\caption{Input representations of different methods on the S3DIS dataset.}\label{tab:s3dis_seg_inputs}
\vspace{-2mm}
\begin{tabular}{l|c|c|c|c|c}
  \hline
 \multirow{2}{*}{Methods} & \multicolumn{3}{c|}{Neural Element}  & \multicolumn{2}{c}{Input Features} \\
 \cline{2-6}
 &Voxel&Point&Super-Point& Geometric & Texture \\
  \hline
  SPG~[{\color{green}30]} & -- & -- & \cmark & [position, observation, geometrics]&\multirow{2}{*}{$[r,g,b]$}\\
SSP+SPG~[{\color{green}29]} & -- & -- & \cmark &[position, radiometry]&\\
\hline
  SEGCloud~[{\color{green}58]} & \cmark & -- & -- & occupancy ${\bf 1}_o$ & \\
 PointNet~[{\color{green}43]} & -- & \cmark & -- & $[x,y,z]$ & \multirow{9}{*}{$[r,g,b]$}\\
Tangent-Conv~[{\color{green}57]} & --  & \cmark& -- & [distance to tangent plane, height, \text{normal}] & \\
PointCNN~[{\color{green}35]}  & -- & \cmark& -- & $[x,y,z,\text{normal}]$ & \\
GACNet~[{\color{green}62]}  & -- & \cmark& -- & [height,~eigenvalues] & \\
SPH3D-GCN~[{\color{green}33]}  & -- & \cmark& -- & $[x,y,z]$ & \\
KPConv~[{\color{green}59]}  & -- & \cmark& -- & $[1,x,y,z]$ & \\
SegGCN~[{\color{green}32]}  & --& \cmark & -- & $[x,y,z]$ & \\
DCM-Net~[{\color{green}52]}  & -- & \cmark & -- & $[x,y,z,\text{normal}]$ & \\
\hline
PicassoNet (Proposed) & -- &\cmark & -- & $[x,y,z]$ & $[r,g,b]$\\
\hline
\end{tabular}
\end{table*}
{\small
[1a] Francis Engelmann, Theodora Kontogianni, Alexander Hermans, and Bastian Leibe.  Exploring spatial context for 3D semantic seg-mentation of point clouds. In Proceedings of the IEEE Conference on Computer Vision and Pattern Recognition, pages 716–724, 2017. 

\noindent[2a] Guohao Li, Matthias M\"{u}ller, Ali Thabet, and Bernard Ghanem.  DeepGCNs:  Can GCNs go as deep as CNNs? In Proceedings of the IEEE International Conference on Computer Vision, 2019.
}
\end{document}